\documentclass[10pt,twocolumn,letterpaper]{article}

\usepackage[pagenumbers]{iccv} %

\usepackage{xcolor}
\usepackage{multirow}

\newcommand{\methodname}{FlashDepth}

\usepackage{pifont}  %
\usepackage{colortbl} %

\definecolor{iccvblue}{rgb}{0.21,0.49,0.74}
\usepackage[pagebackref,breaklinks,colorlinks,allcolors=iccvblue]{hyperref}
\usepackage{microtype}

\def\paperID{9113} %
\def\confName{ICCV}
\def\confYear{2025}

\title{FlashDepth: Real-time Streaming Video Depth Estimation at 2K Resolution}

\author{
    Gene Chou\textsuperscript{1,2} \quad
    Wenqi Xian\textsuperscript{1} \quad
    Guandao Yang\textsuperscript{3} \quad
    Mohamed Abdelfattah\textsuperscript{2}\\[2pt]
    Bharath Hariharan\textsuperscript{2} \quad
    Noah Snavely\textsuperscript{2} \quad
    Ning Yu\textsuperscript{1} \quad
    Paul Debevec\textsuperscript{1}\\[8pt]
    \textsuperscript{1}Netflix Eyeline Studios \quad
    \textsuperscript{2}Cornell University \quad
    \textsuperscript{3}Stanford University 
}

\begin{document}

\twocolumn[{
\maketitle
\begin{center}
\includegraphics[width=0.99\textwidth]{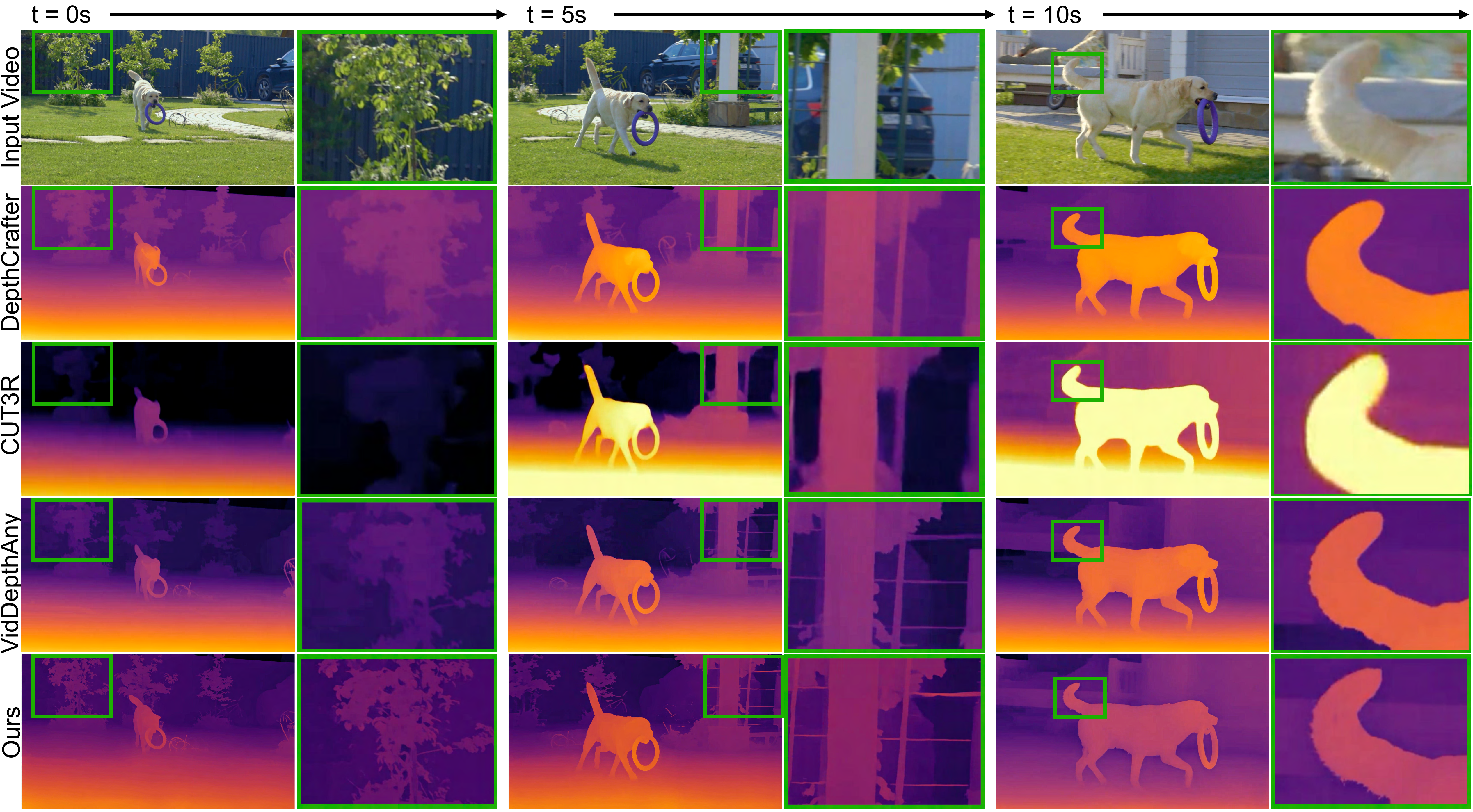}
\captionof{figure}{We present FlashDepth, a video depth estimation model that processes high-resolution streaming videos in real-time. 
Given an in-the-wild video shown above (left column: original input and output; right column: zoomed-in), our method processes it in a streaming fashion at \textbf{2044$\times$1148} resolution at \textbf{24 FPS}. In contrast, the state-of-the-art video depth model, DepthCrafter~\cite{DepthCrafter}, runs at 2.1 FPS at 1024$\times$576 resolution and requires optimizing 110 images at once. Concurrent works like CUT3R~\cite{cut3r} supports streaming inputs (512$\times$288 at 14 FPS) but produces blurry depth; Video Depth Anything (VidDepthAny)~\cite{video_depth_anything} runs at 24 FPS at 924$\times$518 resolution but requires optimizing 32 images at once. Our predicted depth maps are noticeably sharper than all baselines, capturing thin structures such as tree leaves and fur. Zoom in for best viewing results.
    }
    \label{fig:teaser}
\end{center}

}]

\let\oldthefootnote\thefootnote
\let\thefootnote\relax
\footnotetext{* This work was conducted during Gene's internship at Netflix Eyeline Studios.}
\let\thefootnote\oldthefootnote

\begin{abstract}
A versatile video depth estimation model should (1) be accurate and consistent across frames, (2) produce high-resolution depth maps, and (3) support real-time streaming. 
We propose \methodname, a method that satisfies all three requirements, performing depth estimation on a 2044\(\times\)1148 streaming video at 24 FPS. 
We show that, with careful modifications to pretrained single-image depth models, these capabilities are enabled with relatively little data and training. 
We evaluate our approach across multiple unseen datasets against state-of-the-art depth models, and find that ours outperforms them in terms of boundary sharpness and speed by a significant margin, while maintaining competitive accuracy. 
We hope our model will enable various applications that require high-resolution depth, such as video editing, and online decision-making, such as robotics.
We release all code and model weights at \href{https://github.com/Eyeline-Research/FlashDepth}{https://github.com/Eyeline-Research/FlashDepth}.
\end{abstract}

\section{Introduction}
\label{sec:intro}

Video depth estimation is a cornerstone for applications such as robotics~\cite{dong2021realtime, liu2019neural}, augmented reality~\cite{ganj2024mobile,watson2023virtual}, and video editing~\cite{gu2025das, luo2020consistent}.
Robots need real-time sensor data to make online decisions, augmented reality devices must provide streaming, consistent depth, and video editing such as object insertion requires high-resolution and precise depth maps.
As such, we identify three key characteristics 
of a video depth estimation model that can support a wide range of applications: 1) consistency and accuracy across a video sequence, 2) the ability to preserve high-resolution and sharp features, and 3) real-time streaming.

Existing methods cannot satisfy all three requirements.
The most consistent and accurate video depth models rely on post-optimization~\cite{nvds, luo2020consistent} or expensive video diffusion models~\cite{DepthCrafter, chronodepth, kuang2024buffer}. They are slow and can be computationally prohibitive to scale to higher resolutions.

Methods that support real-time and streaming inputs are inaccurate because they rely on lightweight models that have limited network capacity~\cite{reccurent_depth_2020, Zhang_2019_ICCV, li2021temporally, khan2023tcod}, or are inconsistent because they cannot share contextual information between frames~\cite{dpt, depth_anything_v2, depth_pro}. These methods also become at least 3.5x slower~\cite{depth_anything_v2, depth_pro} than real-time 
at 2K resolution due to computational bottlenecks.

\begin{figure*}[t!]
    \centering 
    \includegraphics[width=1\textwidth]{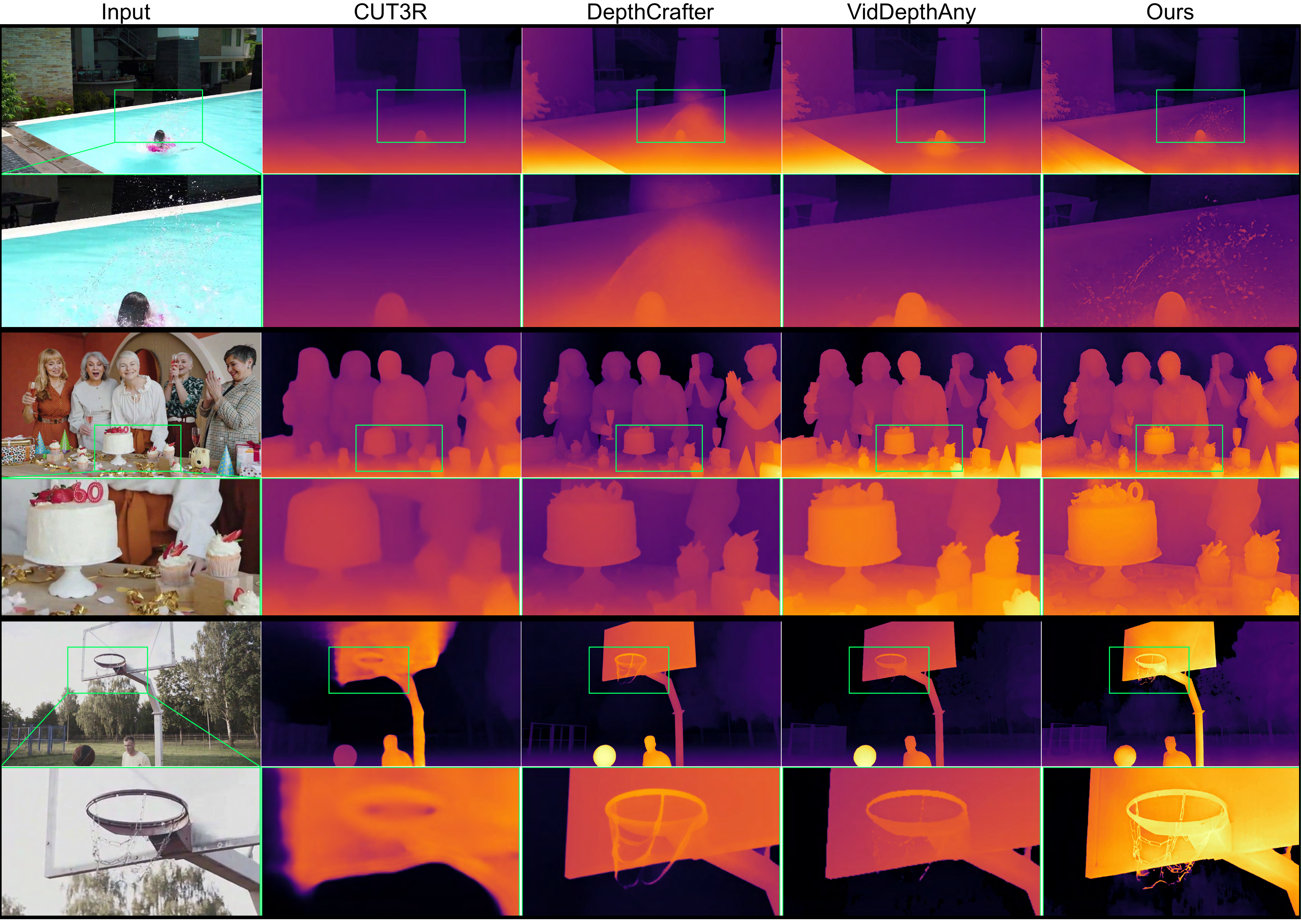}
    \caption{We visualize predicted depth maps (top row) of each method on in-the-wild videos. We also zoom-in (green bounding boxes) to emphasize the boundaries (bottom row). Our method processes high-resolution frames natively and produces sharper boundaries, especially for thin structures and small or far objects.
    }
    \label{fig:results-wild}
    \vspace{-1em}
\end{figure*}

In this paper, we present \methodname{}, a model that 
satisfies all three requirements.
Given a streaming video input at 2K resolution,\footnote{We refer to 2K resolution as the long side containing over 2000 pixels. For instance, Digital Cinema Initiatives defines a 2K format with a resolution of 2048$\times$1080. Our method runs at 24 FPS up to 2044$\times$1148 resolution (patch size 14, aspect ratio 1.78) on an A100 GPU.} it processes all pixels at 24 FPS on an A100 GPU. 
It produces depth maps with high accuracy and sharp boundaries across multiple unseen datasets and in-the-wild videos, shown in \cref{fig:teaser} and \cref{fig:results-wild}.

We achieve this by building on top of a state-of-the-art single-image depth model, Depth Anything v2 (DAv2)~\cite{depth_anything_v2}, rather than finetuning video diffusion models~\cite{DepthCrafter, chronodepth}. While the latter approach achieves consistency for free, DAv2 is meticulously trained on large amounts of synthetic and real data and across multiple stages with multiple losses~\cite{midas, depth_anything_v1, depth_anything_v2}. 
Its per-frame depth accuracy is significantly higher, while maintaining a lightweight architecture that is essential for real-time performance.

We enhance DAv2 with two key technical contributions. First, we add a recurrent neural network that aligns intermediate depth features of individual frames to the same scale, on-the-fly. An accurate relative depth map differs from the ground truth only by a scale and shift factor, so determining the correct scaling factors for each individual depth map aligns them to the same scale. We hypothesize that a lightweight recurrent network is sufficient for this task, but we align the intermediate features, which are computationally cheaper to handle, rather than the final depth maps. We empirically find that this simple approach generalizes to long sequences, moving objects, and large camera pose changes.

Second, we design a hybrid model that balances quality and efficiency. The runtime of DAv2 scales with resolution;
it processes 2K frames at a slow 6 FPS. Our hybrid model features a lightweight component that processes the full 2K resolution to preserve sharp boundaries, but fuses more accurate features from a computationally intensive component that runs at lower resolution.
This approach is motivated by the observation that the raw accuracy of depth maps (i.e., estimating the correct depth relationships between pixels) is less dependent on resolution because boundaries that separate foreground and background account for less than 1\% of pixels, on average, but getting sharp boundaries requires high-resolution processing.

We validate our method and design choices through extensive experiments. We will release all code and model weights. 
In short, we make the following findings and contributions:
\begin{enumerate}
    \item We propose a real-time method to estimate depth for streaming videos at 2K resolution, supporting applications that require high-resolution and realism.
    \item We compare our method to the state-of-the-art and find that it achieves competitive accuracy across multiple datasets, even compared to methods that process large batches of images at once (as opposed to streaming).
    \item We build on open-source pretrained models and require relatively little additional data, making our method easy to train and circumventing the shortage of high-quality video depth data.
\end{enumerate}

\section{Related Work}
\label{sec:relatedwork}

\paragraph{Depth Estimation from a Single Image.}
Recent advances in single-image depth models~\cite{dpt, midas, depth_anything_v1, depth_anything_v2, marigold, lotus, depth_pro, yin2021learning, fu2024geowizard, yin2023metric, hu2024metric3dv2} have been enabled by large amounts of data, both real~\cite{MegaDepthLi18, tung2024megascenes, baruch2021arkitscenes, eth3d} and synthetic~\cite{roberts2021hypersim, wang2020tartanair, cabon2020vkitti2}, and by scalable pretrained architectures, such as vision transformers (ViTs)~\cite{dosovitskiy2020vit, oquab2023dinov2} and diffusion models~\cite{ho2020denoising, rombach2022high}. 
However, single-image models produce flickering, inconsistent depth when applied per-frame to a video sequence. This is true even for metric depth prediction models because of the inherent ambiguity given only a single image.

\paragraph{Optimization-based Video Depth Estimation.} Another common approach is to optimize depth for multiple images jointly. Such methods include classic and learnable dense visual SLAM
methods~\cite{teed2021droidslam, yang2020d3vo, wang2023vggsfm}, as well as methods that optimize the outputs of single-image models to a global scale~\cite{nvds, zhang2021consistent, luo2020consistent}. 
BTimer~\cite{liang2024btimer} produces a reconstruction in a single forward pass, but requires the entire video as input. 
Prompt Depth Anything~\cite{lin2024promptda} 
aligns relative depth to sparse, low-resolution LiDAR to produce 4K depth maps but relies on external sensors.

Another line of recent work estimates pointmaps (e.g., DUSt3R~\cite{dust3r_cvpr24}), from which depth maps and camera parameters can be computed. 
Follow-up work shows that these methods can generalize to videos and dynamic scenes~\cite{monst3r, align3r, murai2024_mast3rslam}, but they still require a global bundle adjustment optimization.
We compare to state-of-the-art pointmap estimation method, and find that they achieve lower depth accuracy and sharpness compared to dedicated depth prediction models. 

Optimization-based methods can produce consistent and accurate depths across a sequence, but are either slow or can only approach real-time speeds at low-resolution (e.g., DROID-SLAM~\cite{teed2021droidslam} averages 15 FPS at 240$\times$320).

\paragraph{Online Video Depth Estimation.} To the best of our knowledge, there are relatively few methods that perform real-time online depth estimation (i.e., operate when only one frame is given at a time). \cite{Zhang_2019_ICCV, reccurent_depth_2020} attempt this setting with LSTMs~\cite{hochreiter1997lstm}. 
\cite{khan2023tcod} stores a global point cloud to maintain consistency. 
\cite{li2021temporally} predicts stereo and motion cues. However, these methods cannot scale to high-resolution frames without substantial changes to their architecture. 
In addition, these methods have only been trained and tested on small datasets and produce depth maps with low accuracy and high flicker~\cite{nvds}.

Aside from the architectural limitations, a likely reason for the poor performance of prior methods is a shortage of video data with moving objects, diverse camera poses, and ground truth depth. 
Among the limited options, self-driving datasets~\cite{waymo, kitti} contain sparse depth with small changes in camera motion, while synthetic scenes with moving objects have limited scene diversity~\cite{zheng2023pointodyssey, karaev2023dynamicstereo}.

Although our approach to enforcing consistency is similar to this class of methods (i.e., we also use a recurrent network to perform online alignment), we revisit this problem setting with modern architectures and training approaches. Our design circumvents the shortage of data and outperforms many state-of-the-art baselines.

\paragraph{Video Depth Estimation using Diffusion Models.} At the time of writing, most state-of-the-art video depth models~\cite{DepthCrafter, chronodepth, yang2024depthanyvideo} finetune video diffusion models~\cite{sora, blattmann2023stable, yang2024cogvideox, chou2024kfcw, ho2022imagen} due to their strong consistency and motion priors, compensating for the lack of abundant, high-quality data.
These methods strike a balance between optimization-based and online approaches because they generally do not take the full video as input but still must process batches of frames, followed by stitching or optimization to align each batch. 
While these methods demonstrate impressive generalization and accuracy, it is non-trivial to scale them to real-time, high-resolution depth due to the compute demands of diffusion models at inference-time.

\paragraph{Concurrent Work.} Video Depth Anything~\cite{video_depth_anything} and CUT3R~\cite{cut3r} are two relevant concurrent works.
Like our method, Video Depth Anything finetunes Depth Anything v2~\cite{depth_anything_v2} on video data with a temporal module. However, their method requires processing batches of 32 images at once, so it does not support streaming and becomes computationally prohibitive to scale in resolution. 
CUT3R can handle streaming inputs, but depth maps produced from their pointmaps are of lower quality than those of a dedicated depth estimation model.
We compare our method to both of these works in \cref{sec:exp}.

\begin{figure}[t!]
    \centering 
    \includegraphics[width=0.46\textwidth]{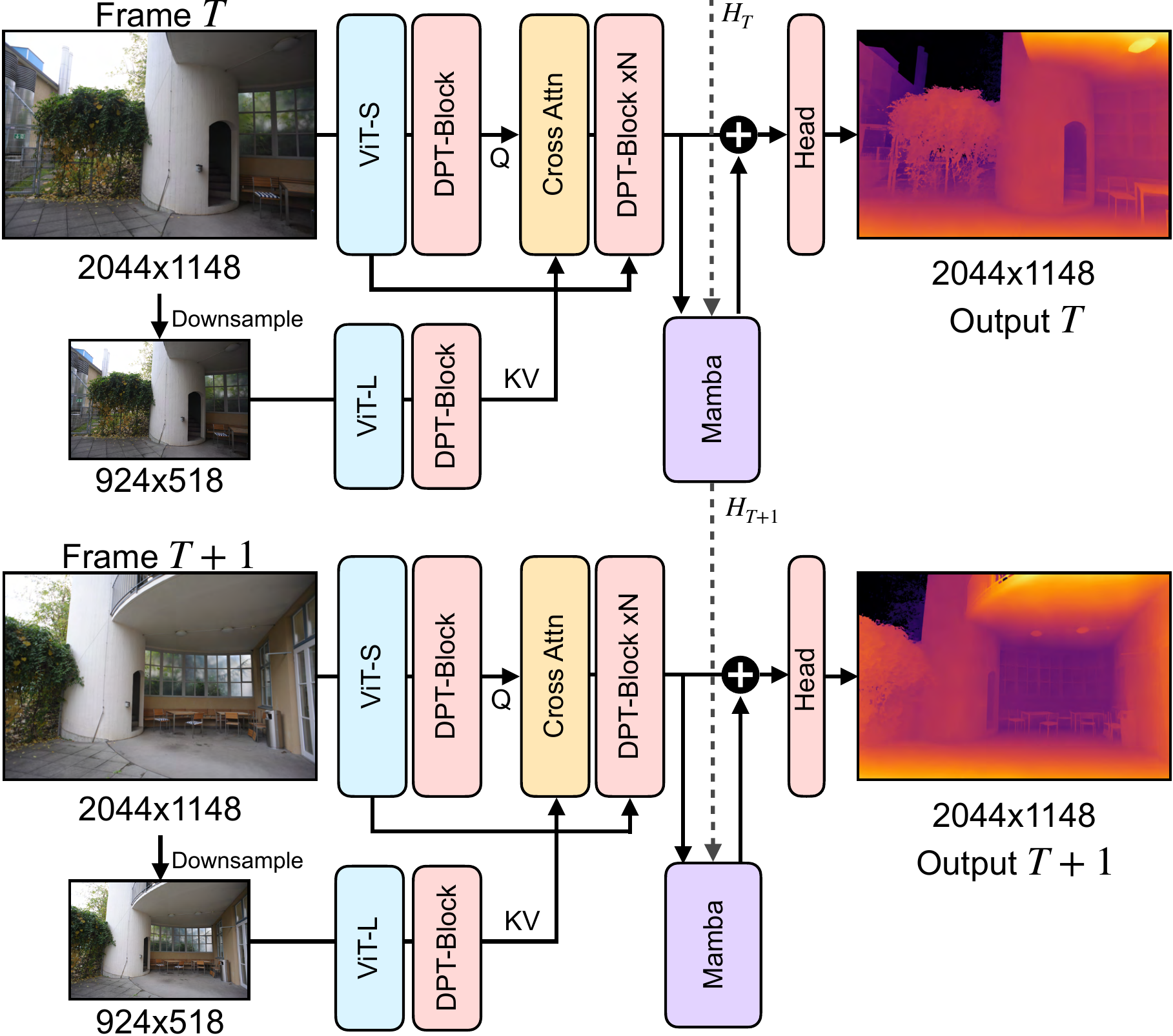}
    \caption{Our architecture builds on DPT~\cite{dpt} and Depth Anything v2~\cite{depth_anything_v2}, but is designed for 
    streaming videos. 
    A recurrent network (Mamba~\cite{mamba2}) aligns frames to a consistent scale on-the-fly, and our hybrid model scales to 2K resolution while maintaining accuracy and throughput of 24 FPS.
    }
    \label{fig:method}
    \vspace{-1em}
\end{figure}

\section{Method}
\label{sec:method}

\methodname{} builds on a pretrained single-image depth model to estimate accurate depth for each individual frame, and enforces consistency for a streaming video in an autoregressive manner. 
In \cref{sec:depthanything}, we 
describe Depth Anything v2~\cite{depth_anything_v2}, our base model. In \cref{sec:mamba} we explain our approach to enforcing temporal consistency.
In \cref{sec:hybrid}, we explain our strategy for training a hybrid model to handle 2K resolution videos while maintaining throughput and accuracy. 
We provide a diagram of our architecture in \cref{fig:method}.

\subsection{Single-Image Depth Base Model}
\label{sec:depthanything}
We use
Depth Anything v2 (DAv2)~\cite{depth_anything_v2} as our single-image base model, which consists of a pretrained DINOv2 ViT~\cite{oquab2023dinov2, dosovitskiy2020vit} encoder and a DPT decoder~\cite{midas, dpt}. 
An image is tokenized and passed through the ViT for feature extraction. 
Four intermediate features layers from the ViT, specifically the outputs of layers 5, 12, 18, and 24 from a 24-layer model, are collected and passed to the DPT decoder.
The motivation is that shallow layers contain more spatial details, while deep layers contain more semantic information. 
The DPT decoder fuses these features using residual convolutional units~\cite{he2016deep} to produce a depth map. 
DAv2 can produce highly detailed and accurate depth maps, but running it on a sequence of frames leads to flickering because there is no common scale.

\subsection{Enforcing Temporal Consistency}
\label{sec:mamba}

To maintain consistent depths across frames, our approach is driven by three considerations. 
First, the model should handle streaming inputs, which means it only has access to the current frame and past information. 
One way to achieve this is using a recurrent network (e.g., an RNN).
The recurrent network maintains a hidden state that encodes information from past frames and updates this hidden state when a new frame arrives.
Specifically, we use Mamba~\cite{gu2023mamba, mamba2} because it has been shown to 
selectively retain relevant information over long sequences.
Our Mamba module processes features from the DPT decoder, and aligns the features while storing context.

Second, we want the scale alignment to incur negligible overhead. 
We assume dense pixel-level features are unnecessary for aligning depths across frames, so we downsample the DPT decoder’s feature output before processing it with Mamba, reducing computational load. 
The alignment operation can be written as
\begin{gather}\label{eq:mamba}
f, H_t = \text{Mamba}(\text{Flatten}(\operatorname{Down}(F)), H_{t-1}) \\
F_{\text{align}} = F + \operatorname{Up}(\text{UnFlatten}(f)),
\end{gather}
where $F$ is the feature output from the DPT decoder with shape $(B,C,H,W)$, $H_t$ is Mamba’s hidden state at time $t$, Flatten reshapes a 4D tensor from $(B,C,H,W)$ to $(B,H\times W,C)$, UnFlatten reverses Flatten, and Down and Up indicate bilinear downsampling and upsampling operations, respectively. 
$F_{\text{align}}$ is then passed through the final convolution head to produce a depth map.
Our temporal module consists of roughly 1\% of the total parameters of our model.
It adds negligible overhead in practice due to its small size and the input's reduced context length after downsampling.

Third, we want to initialize our temporal module such that it maintains the performance of the base single-image depth model.
To achieve this, we place the Mamba model after the DPT decoder, only before the final convolution head, to preserve the integrity of the pretrained features. 
We zero-initialize the output layer of the Mamba model so that the first iteration of \methodname{} remains a valid depth map. 
This preserves the original capabilities of the model at initialization while allowing the training procedure to gradually nudge it toward our consistency objective.

In summary, our pipeline processes each input image individually through the ViT encoder and DPT decoder to produce a feature map. 
Subsequently, the Mamba module adjusts each feature map and updates its hidden state, given the features of the current frame. 
Finally, the adjusted features pass through the convolution head to generate consistent depth predictions. 
This pipeline is illustrated in \cref{fig:method} (so far, we have described only the high-resolution path through ViT-S).

\subsection{Hybrid Model for Efficiency at High-Resolution}
\label{sec:hybrid}

While our temporal module adds negligible latency, the original model (DAv2-Large) achieves only 6 FPS at 2K resolution.
It is difficult to simultaneously scale up both model size and resolution while achieving real-time performance. 
To address this, we examine the runtime and accuracy trade-off across the existing DAv2 family~\cite{depth_anything_v1, depth_anything_v2}, which includes three checkpoints---DAv2-Small (S), DAv2-Base (B), and DAv2-Large (L)---finetuned from DINOv2’s ViT-Small, ViT-Base, and ViT-Large backbones, respectively. 
DAv2-L achieves the highest accuracy across all benchmarks but is slow at high resolution. 
Conversely, DAv2-S achieves real-time throughput at 2K resolution but yields a substantial accuracy drop, so directly reducing the model size is an inadequate solution.

To achieve real-time depth estimation at 2K resolution without compromising accuracy, we propose a hybrid model that processes frames in two streams, based on the following considerations. 
First, for real-time throughput, the main stream that processes 2K frames must be lightweight. We denote this stream \methodname-S, finetuned from DAv2-S and adding our temporal module.

Second, to maintain accuracy, we incorporate the features from a more powerful model, namely DAv2-L. We train a second stream \methodname-L that finetunes DAv2-L and processes frames at a lower resolution (518px) to produce accurate features that can be used to supervise the intermediate features of \methodname-S. 
We use cross-attention~\cite{vaswani2017transformer} to fuse the intermediate features without modifying the original architectures. 

We observe that while resolution is crucial for preserving sharp boundaries, it has minimal impact on numerical accuracy, since boundaries account for less than 1\% of pixels, on average. For instance, we test the per-frame accuracy of DAv2 on multiple datasets at 2K vs 518px (resizing short side to 518 while maintaining aspect ratio) resolution, and find that the difference in accuracy between the two is negligible. However, we start seeing degradation in accuracy at even lower resolutions.

We show this pipeline in \cref{fig:method}. A high-resolution image (2044$\times$1148) is passed through \methodname-S. 
Concurrently, a downsampled image (924$\times$518) is passed through \methodname-L. 
The intermediate features from the larger model are fused with those of the smaller model:
\begin{equation}\label{eq:crossattn}
    F_{\text{fused}_i} = \text{CrossAttn}(Q=F_{\text{S}_i}, KV=F_{\text{L}_i}),
\end{equation}
where $F_{\text{S}_i}$ and $F_{\text{L}_i}$ denote intermediate feature representations at the $i$-th layer of the DPT decoders from the small and large models, respectively. 

We zero-initialize the module such that at the first training iteration, $F_{\text{fused}_i} = F_{\text{S}_i}$.
This encourages \methodname-S to gradually learn to use $F_{\text{L}_i}$ without abrupt changes to its learned weights. 
The cross-attention module is lightweight and processes a sequence length of $M\times N$ where $M$ is the sequence length of the high-resolution image features, and $N$ is the sequence length of the base-resolution image features (the output of the first DPT layer maintains the same sequence length as the input to the transformer). 
By training this hybrid model on small amounts of video depth data, it learns to use the cross-attention module to extract the more accurate features, while preserving sharpness and speed.

By running both models concurrently, even with communication between the two, we can process 2K images at 24 FPS (wall time) on an A100 GPU. 
Empirically, we show that our hybrid model's accuracy is comparable to that of \methodname-L (\cref{tab:depth-consistency}), and is noticeably higher than that of \methodname-S (\cref{tab:ablation-hybrid-small}). 

\subsection{Two Stage Training}
\label{sec:training}

Training our model end-to-end requires a large number of videos with high-resolution dense depth maps, but most existing datasets are sub-720p. 
Instead, we train our model in two stages.
First, we use low-resolution depth data to train \methodname-L and \methodname-S (\cref{sec:mamba}) to be consistent across video sequences.
Then, we use a small amount of high-resolution data to supervise a hybrid model (\cref{sec:hybrid}) initialized from \methodname-L and \methodname-S.

In the first stage, we initialize the ViT encoders and DPT decoders of \methodname-L and \methodname-S from DAv2-L and DAv2-S, respectively, and train the temporal modules from scratch.
We use the following datasets: MVS-Synth~\cite{DeepMVS}, Spring~\cite{spring_dataset}, TartanAir~\cite{wang2020tartanair}, PointOdyssey~\cite{zheng2023pointodyssey}, and Dynamic Replica~\cite{karaev2023dynamicstereo}. After filtering out invalid scenes, we obtain roughly half a million image-depth pairs. We process all frames to a fixed resolution of $518\times 518$ via random cropping. 
Since these datasets are synthetic and metric, we supervise our model using a simple L1 loss, instead of the scale-shift-invariant loss used by many prior works~\cite{depth_anything_v2, midas}. 
We do not use any additional supervision, such as optical flow~\cite{monst3r, align3r, kuang2024buffer} or temporal losses~\cite{khan2023tcod, reccurent_depth_2020}.

Due to memory constraints, we perform backpropagation with input sequences containing a small number of frames, but augment the dataset by setting longer strides between frames.
This allows our model to maintain consistency when tested on videos containing up to 1000 frames, and even outperform state-of-the-art video depth models~\cite{DepthCrafter} that perform an explicit optimization designed to stitch shorter sequences into longer ones.

In the second stage,
we train our hybrid model using two high-resolution depth datasets with 2K depth annotations: MVS-Synth~\cite{DeepMVS} and Spring~\cite{spring_dataset}, which contains a total of roughly 16,000 image-depth pairs.
We freeze the weights in \methodname-L and only finetune \methodname-S.
\methodname-S processes images at their original 2K resolution, and \methodname-L processes the downsampled version (short side 518) of the input images.
We initialize the cross-attention blocks that transfer features between the DPT decoders from scratch. 
Again, we supervise this model using an L1 loss. 

As shown in \cref{fig:method}, only the intermediate features from DPT, rather than from Mamba, are taken from \methodname-L, but training \methodname-L in the first stage is crucial for aligning the features of \methodname-L and \methodname-S to roughly the same scale.

\subsection{Implementation Details}
\label{sec:implementation}

Each of our Mamba modules~\cref{eq:mamba} consists of four blocks.
Each block contains layer normalization, a Mamba layer, and an MLP. 
Empirically, we found that a vanilla Mamba implementation, rather than a bidirectional scan~\cite{vim}, is sufficient for our task.

The CrossAttn module in \cref{eq:crossattn} consists of two transformer blocks with cross-attention. It takes queries (Q) from the high-resolution features ($F_\text{S}$), and keys and values (KV) from the low-resolution, but more accurate features ($F_\text{L}$). We only fuse the output features of the first DPT layers to avoid low resolution artifacts seeping into the high-resolution stream.

During first-stage training, we set the Mamba learning rates to 1e-4, and the rest of the model to 1e-6. 
Each takes half a day on 8 A100 GPUs. During the second stage, we freeze \methodname-L, set the cross-attention module's learning rate to 1e-4, and finetune the \methodname-S model with learning rate 1e-6. This also takes half a day on 8 A100 GPUs. 

During inference, we load one image at a time. Since this does not fully utilize the GPU, we can create CUDA graphs to run the two streams (up to the cross-attention module) in parallel. Additionally, our method defaults to running \methodname-L directly when the short side of the image has fewer than 518 pixels, since it is capable of running in real-time at that resolution.

To facilitate reproduction, we will release all code and model weights.

\begin{figure*}[t!]
    \centering 
    \includegraphics[width=0.99\textwidth]{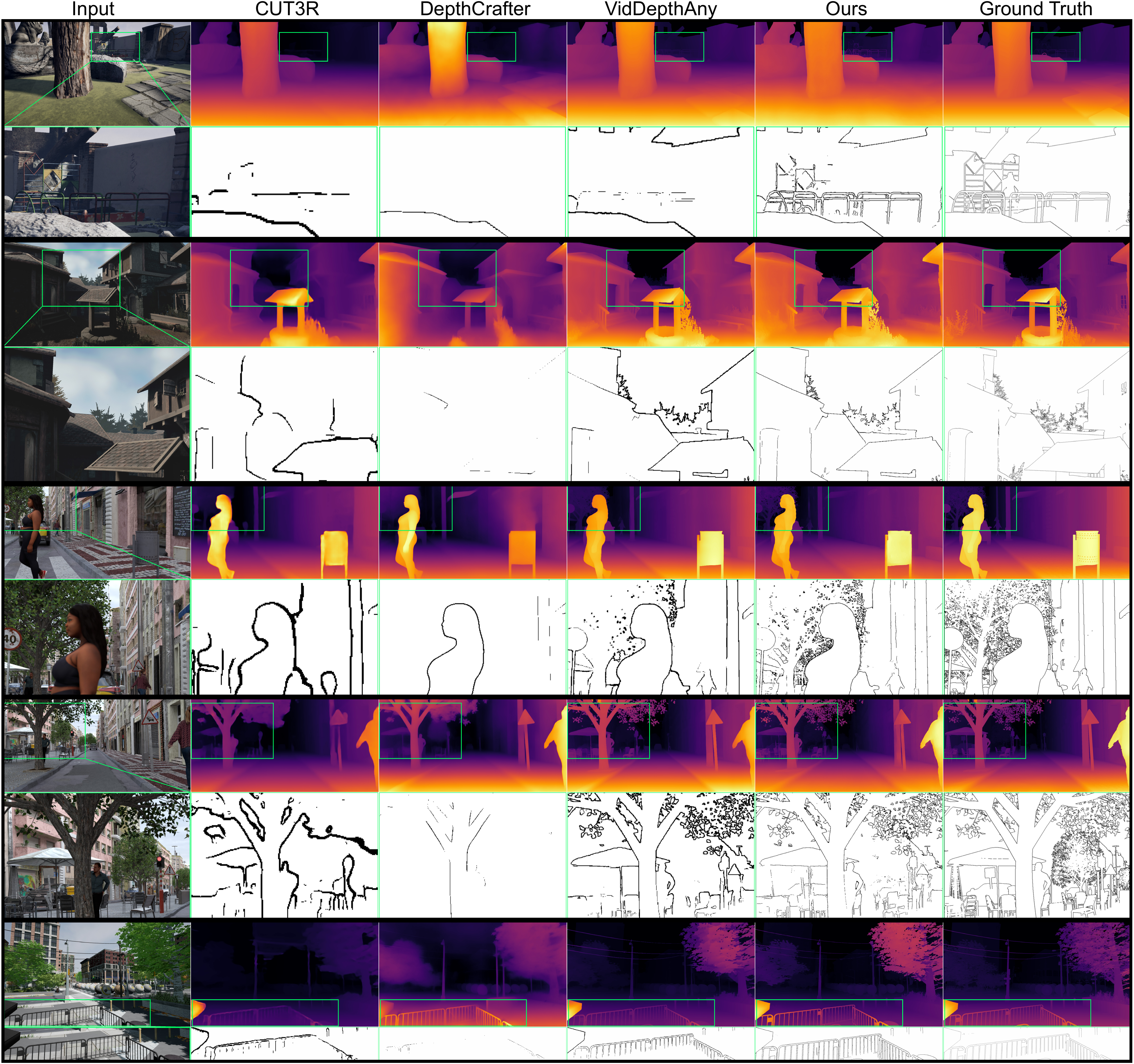}
    \caption{We provide more results of predicted depth maps (top row) on high-quality synthetic datasets (UnrealStereo4K~\cite{unreal4k} and UrbanSyn~\cite{urbansyn}) and zoom-in (green bounding boxes) to show the edge maps (bottom row), which are defined by a depth value change between neighboring pixels, above a threshold. This allows us to quantitatively evaluate the predicted boundaries with ground truth depth (\cref{sec:resolution}). Compared to baselines, our method captures thin structures and small or far objects.
    }
    \label{fig:results}
    \vspace{-1em}
\end{figure*}

\section{Experiments}
\label{sec:exp}

\paragraph{Baselines.} We compare our method to DepthCrafter~\cite{DepthCrafter}, Video Depth Anything~\cite{video_depth_anything}, CUT3R~\cite{cut3r}, and Depth Anything v2~\cite{depth_anything_v2}.
DepthCrafter~\cite{DepthCrafter} is finetuned from a video diffusion model and demonstrates strong generalization 
and higher accuracy compared to previous video depth models, such as ChronoDepth~\cite{chronodepth} and NVDS~\cite{nvds}. 
Video Depth Anything (VidDepthAny)~\cite{video_depth_anything} is concurrent work that finetunes Depth Anything v2 (DAv2)~\cite{depth_anything_v2} with a temporal module. For all experiments, we use their largest model.
CUT3R~\cite{cut3r} is also concurrent work that estimates pointmaps from streaming videos by designing a transformer-based recurrent model. 
We include DAv2~\cite{depth_anything_v2} because it is the base model of VidDepthAny and our method, and serves as a reference point for understanding the effects of our approaches. For all experiments, we use their DAv2-Large model.

\paragraph{Metrics.}
For each scene, we denote with $\hat{d}$ the predicted depths across the sequence with shape $(B,h,w)$ where $B$ is the number of frames in the scene, and $d$ the ground truth depth, with shape $(B,H,W)$. Following standard practice, we run each method on its default resolution ($h,w$), which is either the default resolution provided in their code, or the resolution it was trained on, while maintaining aspect ratio. 
Then we bilinear interpolate the outputs to $H,W$ and use least squares to find a scale and shift factor to align the entire predicted sequence with ground truth (as opposed to finding a per-frame scale and shift).

We use three metrics. The first two measure accuracy ($\delta_1 \uparrow$) and relative error (AbsRel $\downarrow$), defined as
$\delta_1 = \text{percentage of pixels where} \max(d/\hat{d}, \hat{d}/d) < 1.25)$ and $\text{AbsRel} = |\hat{d}-d|/d$, respectively.
The third is the boundary metric proposed in \cite{depth_pro}: if the ground truth neighboring pixels differ by a threshold, we classify it as an boundary, and we calculate the F1 score to determine whether the predicted depths have the same classification. We provide more details in the supplement. This metric focuses on fine details and sharp boundaries, which tend not to be measured in error and accuracy metrics because the boundaries represent just 1\% of pixels, on average. We show depth maps and their corresponding boundary maps in \cref{fig:results-wild} and \cref{fig:results}.

\paragraph{Datasets.}
We test on five diverse datasets. ETH3D~\cite{eth3d} contains 13 real scenes, both outdoor and indoor. Sintel~\cite{sintel} is a 3D animated film with many moving objects. Waymo~\cite{waymo} is a self-driving dataset with continuous camera motion. We randomly sampled 30 scenes from its validation set. Unreal4K~\cite{unreal4k} and UrbanSyn~\cite{urbansyn} are both synthetic but photorealistic datasets with diverse camera viewpoints and scenes. 
Unreal4k has nine scenes in total, whereas UrbanSyn contains a total of more than 7,000 frames. 
We take the first 1,000 frames of UrbanSyn as our testing data.
Even though the images in ETH3D, Unreal4k, and UrbanSyn are not organized as continuous video sequences, we stitch these images into a video to mimic video frames with significant camera motion. 
We compute both relative error and absolute accuracy metrics in all five datasets.
We report boundary sharpness using only Unreal4k and UrbanSyn, since only these two contain dense and high-resolution ground truth depth maps.

\begin{table}[t]
    \centering
    \footnotesize
    \setlength{\tabcolsep}{2.5pt}  %
    \renewcommand{\arraystretch}{1.0}  %
    \begin{tabular}{lcccc}  
        \toprule
        Method & Unreal4K~\cite{unreal4k} & UrbanSyn~\cite{urbansyn} & FPS ↑ & Resolution \\
        & (3840\(\times\)2160) & (2048\(\times\)1024) & & (AR 1.77) \\
        & F1 ↑ & F1 ↑ &  &  \\
        \midrule
        DepthAny v2~\cite{depth_anything_v2}  & 0.058 & 0.118 & 30 & 924\(\times\)518 \\
        DepthCrafter$^*$~\cite{DepthCrafter} & 0.021 & 0.044 & 2.1 & 1024\(\times\)576 \\
        VidDepthAny$^*$~\cite{video_depth_anything}& 0.049 & 0.097 & 24 & 924\(\times\)518 \\
        CUT3R~\cite{cut3r} & 0.007 & 0.019 & 14 & 512\(\times\)288 \\
        \midrule
        \methodname-L & 0.048 & 0.136 & 30 & 924\(\times\)518 \\
        \methodname-L (high) & \textbf{0.143} & \textbf{0.271} & 6.0 & 2044\(\times\)1148 \\
        \methodname{} (Full) & \underline{0.109} & \underline{0.185} & 24 & 2044\(\times\)1148 \\
        \bottomrule
    \end{tabular}
    \caption{We report the F1 score of predicted vs ground truth boundaries, defined by a value change between neighboring pixels, across multiple thresholds.
    For all methods, we use their default resolution (i.e., the resolution provided in their demo or used during training) while preserving aspect ratio. We also report the corresponding FPS when run at the denoted resolution, over 1000 frames. Results validate that processing frames natively at high-resolution is crucial for preserving fine details (\methodname-L vs \methodname-L (high)). $^*$ means the method does not support streaming, but process batches of images instead. 
    }
    \label{tab:depth-boundaries}
    \vspace{-1em}
\end{table}

\begin{table*}[t]
    \centering
    \small
    \setlength{\tabcolsep}{3pt}  %
    \renewcommand{\arraystretch}{1.0}  %
    \begin{tabular}{lccccccccccc}  
        \toprule
        Method & \multicolumn{2}{c}{ETH3D (30 frames)~\cite{eth3d} } & \multicolumn{2}{c}{Sintel (50)~\cite{sintel}} & \multicolumn{2}{c}{Waymo (200)~\cite{waymo}} & \multicolumn{2}{c}{Unreal4K (1k)~\cite{unreal4k}} & \multicolumn{2}{c}{UrbanSyn (1k)~\cite{urbansyn}} & $\delta_1$ rank \\
        & AbsRel ↓ & $\delta_1$ ↑ & AbsRel ↓ & $\delta_1$ ↑ & AbsRel ↓ & $\delta_1$ ↑ & AbsRel ↓ & $\delta_1$ ↑ & AbsRel ↓ & $\delta_1$ ↑ & \\
        \midrule
        DepthAny v2~\cite{depth_anything_v2}  & 0.222 & 0.633 & 0.373 & 0.561 & 0.108 & 0.897 & 0.401 & 0.379 & 0.209 & 0.622 & 5.4 \\
        DepthCrafter$^*$~\cite{DepthCrafter} & 0.163 & 0.745 & 0.283 & \textbf{0.697} & 0.143 & 0.790 & 16.52 & 0.399 & 0.266 & 0.556 & 4.6 \\
        VidDepthAny$^*$~\cite{video_depth_anything} & \textbf{0.119} & \underline{0.864} & \textbf{0.264} & \underline{0.660} & \textbf{0.076} & \textbf{0.944} & 0.414 & \underline{0.606} & \underline{0.110} & \textbf{0.892} & \textbf{1.6} \\
        CUT3R~\cite{cut3r} & \underline{0.124} & 0.836 & 0.465 & 0.509 & 0.100 & 0.902 & \textcolor{gray}{\textbf{0.152}} & \textcolor{gray}{\textbf{0.851}}& \textcolor{gray}{0.120} &  \textcolor{gray}{0.878} & 3.6 \\
        \midrule
        \methodname-L & \textbf{0.119} & \textbf{0.875} & \underline{0.265} &  0.642 & \underline{0.098} & \underline{0.924} & \underline{0.276} & 0.566 & \textbf{0.109} & \underline{0.882} & \underline{2.2} \\
        \methodname{} (Full) & 0.132 & 0.848 & \underline{0.265} &  0.642 & 0.101 & 0.916 & 3.373 & 0.545 & 0.126 & 0.862  & 3.4 \\
        \bottomrule
    \end{tabular}
    \caption{We compare error and accuracy metrics across five diverse datasets. The number next to the dataset name is the number of frames in each scene. The predicted depth is aligned with ground truth using a global scale and shift along the sequence to calculate metrics. $^*$ means the method does not support streaming, but process batches of images instead. \textcolor{gray}{Gray} means the method was trained on the test data. Based on the average $\delta_1$ rank, our method's accuracy only trails behind that of Video Depth Anything~\cite{video_depth_anything} (concurrent work), which optimizes batches of 32 images, while ours handles streaming inputs (i.e., one frame at a time).}
    \label{tab:depth-consistency}
    \vspace{-1em}
\end{table*}

\subsection{Depth Sharpness, Resolution, and Runtime}
\label{sec:resolution}

We first evaluate the boundary sharpness and runtime of each method, which is the focus of ours, and crucial for online tasks and applications that require high-resolution. 
We use the F1 score proposed in \cite{depth_pro}. The boundaries are defined by a value difference between neighboring pixels, and we calculate these boundaries across multiple thresholds.
We visualize the boundaries of a fixed threshold as an example in \cref{fig:method}. Our method captures thin structures like poles and leaves and overall produces sharper boundaries compared to existing methods.

In addition to the four baselines, we include results for our full hybrid model (\methodname{} (Full)), large model (\methodname-L), and the large model run at higher resolution large model (\methodname-L (high)). 
We report quantitative results in \cref{tab:depth-boundaries}. Boundary sharpness is correlated to resolution, as a low-resolution image blurs or averages out boundaries. Thus, it is not surprising that \methodname{} (Full) significantly outperforms baselines as it operates at a higher resolution. However, \methodname-L (high) also produces sharper boundaries than \methodname{} (Full), showing that there also exists a tradeoff between model capacity and boundary sharpness, even if resolution seems to be the most important factor. 

Existing video depth models cannot be easily extended to higher resolutions. DepthCrafter and VidDepthAny process 110 and 32 frames, respectively, and run out of memory at higher resolutions. For instance, running VidDepthAny at 924$\times$518 resolution takes up 40gb of vram. CUT3R is only trained on resolutions under 518px (long side 518), and performance degrades at higher resolutions. 

We also report the FPS in wall time at the denoted resolutions. For a scene in Unreal4k with 1000 images, we start the timer after a single warmup iteration, and record the time after all frames are processed. 
From our experiments, only our hybrid approach is capable of operating at high-resolution while maintaining real-time performance of 24 FPS.

\subsection{Depth Error and Accuracy}
\label{sec:accuracy}

We report all error and accuracy metrics in \cref{tab:depth-consistency}. 
All methods are run at the resolution reported in \cref{tab:depth-boundaries}.
While we visualize per-frame depth maps in \cref{fig:method}, we encourage readers to view our supplemental videos to visualize each method's consistency over video sequences.

VidDepthAny achieves the overall highest accuracy across the five datasets, with our base model coming in second. VidDepthAnything processes 32 images at once while ours only sees one frame at a time, but still demonstrates competitive performance.

DepthCrafter is finetuned from a video diffusion model and processes 110 images at once, but our method produces higher accuracy across all but one datasets. Since we align the entire sequence to ground truth using only a global scale and shift for comparison, this shows that our temporal module can align streaming frames to the same scale without an expensive model. 

Our method also performs better than CUT3R, even though CUT3R was trained on 33 datasets. 
This indicates that building on top of a dedicated, pretrained depth estimation model with small amounts of data can significantly simplify training while achieving high performance, compared to training from scratch.

Both VidDepthAny and our method perform better than DAv2 on all datasets, up to a 20\% increase in accuracy in many cases, showing that a temporal alignment (regardless of the implementation) is a promising direction for generalizing single-image models to the video domain, rather than relying on expensive video diffusion models. 

By comparing the two variations of \methodname{}, we verify that our cross-attention module successfully transfers the robust features predicted by the base model. \methodname{} (Full) is ranked third among the methods, and generally trails behind its counterpart \methodname-L by only 1-2\%. Note that, as mentioned in \cref{sec:implementation}, \methodname{} (Full) defaults to running \methodname-L when the short side of the image has less than 518 pixels, so the metrics for Sintel are the same for both methods. 

In contrast, we show in \cref{sec:ablation} that across all datasets, \methodname-S performs noticeably worse, validating that directly decreasing model capacity is not a desirable tradeoff.

\begin{table}[]
    \centering
    \footnotesize
    \renewcommand{\arraystretch}{1.0}  %
    \begin{tabular}{lcccccc}  
        \toprule
        Method & \multicolumn{2}{c}{Unreal4K~\cite{unreal4k}} & \multicolumn{2}{c}{UrbanSyn~\cite{urbansyn}} & FPS ↑ \\
        & \multicolumn{2}{c}{(3840\(\times\)2160)} & \multicolumn{2}{c}{(2048\(\times\)1024)} & \\
        & \(\delta_1 \uparrow\) & F1 ↑ & \(\delta_1 \uparrow\) & F1 ↑ &   \\
        \midrule
        \methodname-L & \textbf{0.566}  & 0.048 & \textbf{0.882}  & \underline{0.136} & 30 \\
        + UNet~\cite{unet} & 0.490 & 0.034 & 0.811 & 0.107 & 24  \\
        + DAGF~\cite{dagf} & 0.377 & \underline{0.055} & 0.703 & 0.133 & 7.7  \\
        \methodname{} (Full) & \underline{0.545} & \textbf{0.109} & \underline{0.862} & \textbf{0.185} & 24  \\
        \bottomrule
    \end{tabular}
    \caption{Ablation study of super-resolution methods. We report the accuracy (\(\delta_1\)) and the F1 score of predicted vs. ground truth boundaries. Using the same base model (\methodname-L), we compare three ways to preserve boundary sharpness: a UNet, a guided-filter super-resolution network (DAGF), and ours. The outputs of all three methods have the same 2K resolution. Our hybrid model (\methodname{} (Full)) produces depth maps that are much sharper than other methods while preserving accuracy on test sets.}
    \label{tab:superres-ablation}
    \vspace{-1em}
\end{table}

\subsection{Ablation Study}
\label{sec:ablation}

\noindent\textbf{Super-Resolution vs Hybrid Model}.
Our hybrid model preserves accuracy and boundary sharpness while maintaining real-time performance. 
But 
can we simply use a dedicated depth super-resolution network?

We examine a recent survey~\cite{zhong2023guided} on depth super-resolution and find that even the most recent methods, whether they use diffusion~\cite{metzger2023guided} or learnable filters~\cite{dagf}, overfit their models to single datasets and their reported numbers show unsatisfactory generalization. Although we did not find ablation studies that trained on multiple datasets, we hypothesize these models are unable to generalize because this requires training with diverse, high-quality data and sometimes multiple stages~\cite{depth_anything_v2, depth_pro}.

We report quantitative results in \cref{tab:superres-ablation}. \methodname-L is our baseline, where we bilinear upsample the 518px outputs to the ground truth resolution. ``+ UNet" is one of our earlier experiments, where we pretrain \methodname-L on our five training datasets at 518px resolution, then freeze it and add a UNet~\cite{unet} trained from scratch. The UNet encodes the high-resolution image, then uses skip connections to upsample the low-resolution depth map features (after our Mamba module and before the final convolution head, so the head is re-trained). We find that this approach hurts both accuracy and boundary sharpness, likely because there is insufficient data to re-train the head to match the performance of the pretrained one. In contrast, our hybrid approach ensures no modifications to the original capabilities of the model during the first training iteration, so it can be finetuned without a lot of data.

We also compare to an off-the-shelf super-resolution network. We apply DAGF~\cite{dagf}, which is a relatively fast method (10 FPS) that uses an attention-guided image filtering technique, to the output depth of \methodname-L (denoted ``+ DAGF"). This led to substantially worse accuracy, dropping by roughly 20\%. In line with the paper's findings, the network cannot generalize to out-of-distribution data. It can only be run after \methodname-L, resulting in an 7.7 FPS.

From \cref{tab:superres-ablation}, \methodname{} (Full) not only preserves boundary sharpness from high-resolution images, but it also maintains accuracy and strong generalization due to the transfer of features from the more powerful \methodname-L. In contrast, neither learning-based nor filtering-based methods match our performance. 

From our observations, super-resolution methods show poor generalization performance, and may require substantial amounts of high-resolution training data to improve results, which are not easily obtainable. Our approach directly leverages pretrained single-image models to avoid this issue.

\begin{table*}[t]
    \centering
    \small
    \setlength{\tabcolsep}{3pt}  %
    \renewcommand{\arraystretch}{1.0}  %
    \begin{tabular}{lcccccccccc}  
        \toprule
        Method & \multicolumn{2}{c}{ETH3D (30 frames)~\cite{eth3d} } & \multicolumn{2}{c}{Sintel (50)~\cite{sintel}} & \multicolumn{2}{c}{Waymo (200)~\cite{waymo}} & \multicolumn{2}{c}{Unreal4K (1k)~\cite{unreal4k}} & \multicolumn{2}{c}{UrbanSyn (1k)~\cite{urbansyn}} \\
        & AbsRel ↓ & $\delta_1$ ↑ & AbsRel ↓ & $\delta_1$ ↑ & AbsRel ↓ & $\delta_1$ ↑ & AbsRel ↓ & $\delta_1$ ↑ & AbsRel ↓ & $\delta_1$ ↑ \\
        \midrule
        \methodname-S & 0.147 & 0.786 & 0.299 & 0.612 & 0.109 & 0.898 & \underline{1.155} & 0.541 & 0.158 & 0.784\\
        \methodname-L & \textbf{0.119} & \textbf{0.875} & \textbf{0.265} &  \textbf{0.642} & \textbf{0.098} & \textbf{0.924} & \textbf{0.276} & \textbf{0.566} & \textbf{0.109} & \textbf{0.882} \\
        \methodname{} (Full) & \underline{0.132} & \underline{0.848} & \textbf{0.265} &  \textbf{0.642} & \underline{0.101} & \underline{0.916} & 3.373 & \underline{0.545} & \underline{0.126} & \underline{0.862}\\
        \bottomrule
    \end{tabular}
    \caption{Ablation study of our hybrid model \methodname{} (Full) vs a smaller model \methodname-S. A smaller model is faster but at the cost of substantially decreased accuracy, as shown in this table: reducing the model size from \methodname-L to \methodname-S leads to a 3-10\% drop. On the other hand, our hybrid model runs at the same speed as \methodname-S, at the cost of a noticeably smaller decrease in accuracy across all datasets.}
    \label{tab:ablation-hybrid-small}
    \vspace{-1em}
\end{table*}

\medskip
\noindent\textbf{FlashDepth (Full) vs FlashDepth-S}. In \cref{sec:hybrid} we discuss the motivation for our hybrid model is that decreasing model size directly leads to a substantial accuracy drop, as shown in the original DAv2~\cite{depth_anything_v2} paper. Here, we run an ablation to verify this observation. In \cref{tab:ablation-hybrid-small}, we compare three variations of \methodname. We find that reducing the model size from \methodname-L ($\sim$300M parameters) to \methodname-S ($\sim$21M parameters) leads to a 3-10\% accuracy drop across all datasets. On the other hand, our hybrid model runs at the same speed as \methodname-S, at the cost of a noticeably smaller decrease (1-3\%). This validates our design and confirms that our cross-attention module transfer more robust features from \methodname-L.

\section{Conclusion}
\label{sec:conclusion}

In this work, we develop a method that performs depth estimation of a 2048$\times$1152 streaming video at 24 FPS. Compared to the state-of-the-art, our method delivers better boundary sharpness while being very competitive in terms of accuracy when processes streaming inputs in real time.

The main limitation of our method is
residual flickering. 
While it may be difficult to match the performance of online methods to that of optimization, it would be possible to concurrently optimize processed frames while streaming new ones for certain applications, similar to SLAM. We leave this for future work.

\section*{Acknowledgments}
Gene Chou was supported by an NSF graduate fellowship (2139899). We thank Chi-Chih Chang for discussions on Mamba and efficiency; Jennifer Lao, Tarik Thompson, and Daniel Heckenberg for their operational support; Nhat Phong Tran, Miles Lauridsen, Oliver Hermann, Oliver Walter, and David Shorey for helping integrate FlashDepth into Eyeline's internal VFX pipeline.

{
    \small
    \bibliographystyle{ieeenat_fullname}
    \bibliography{main}
}

\clearpage
\setcounter{page}{1}
\maketitlesupplementary

\section{Evaluating Temporal Consistency}
To evaluate temporal consistency, we record the optimal scale factor between each individual predicted depth map and its corresponding ground truth frame to measure the drift in scale.
Specifically, we use the Waymo test set, which consists of 200 frames per scene. 

\begin{figure}[t]
    \centering
    \includegraphics[width=0.4\textwidth]{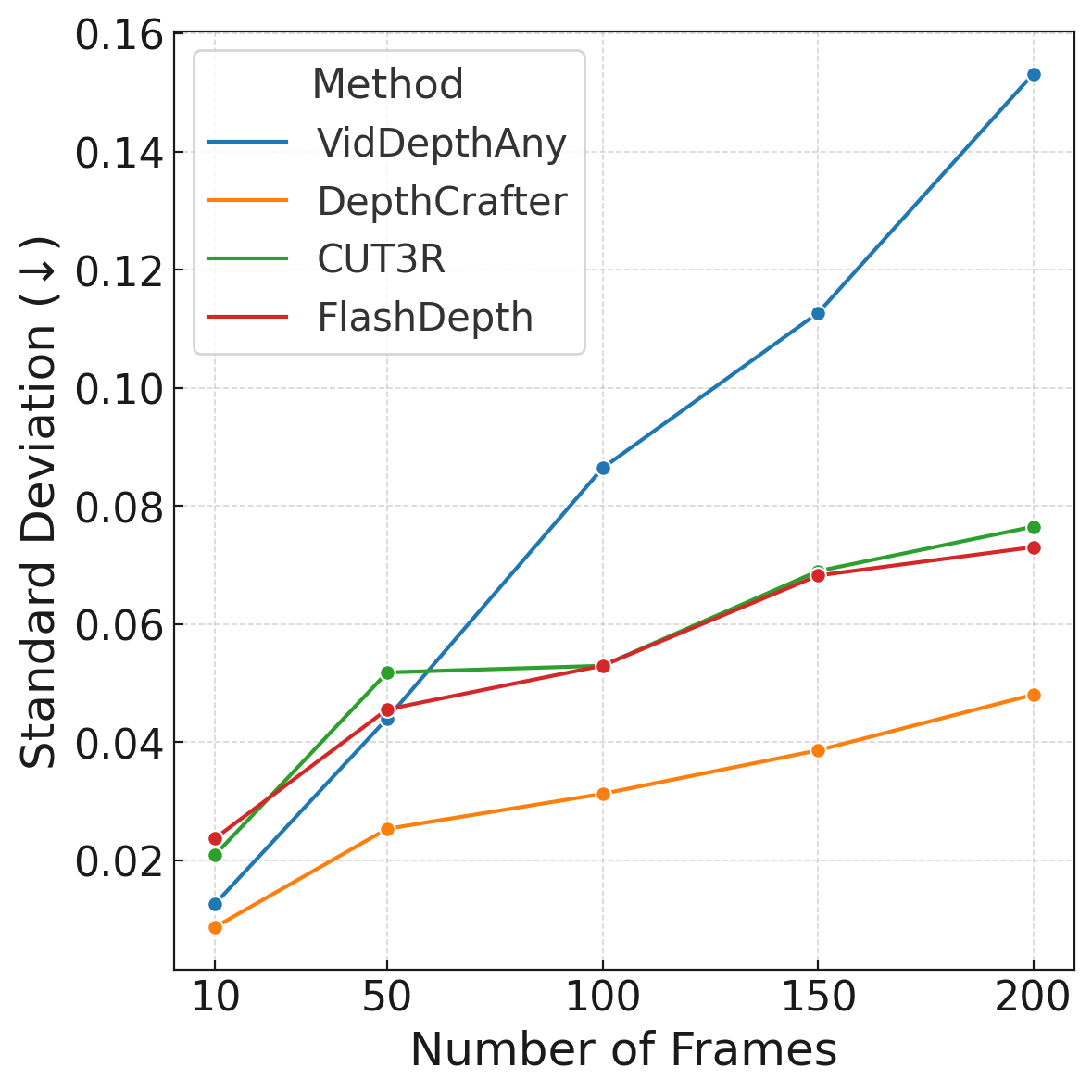}
    \caption{Standard deviations of scale averaged over all Waymo test sequences.}
    \label{fig:temporal-std}
    \vspace{-1em}
\end{figure}

We provide a plot in \cref{fig:temporal-std}, which shows the average standard deviation of scale factors across all test sequences from Waymo, where the frame index on the $x$-axis indicates the standard deviation of scale factors for the first $x$ video frames. 
DepthCrafter yields the lowest numbers and is the most temporally consistent, though it is slow and cannot output high-resolution frames. 
FlashDepth and CUT3R are comparable, likely because the mechanisms for handling new input frames both involve recurrent networks with a hidden state, though FlashDepth is more accurate.
On the other hand, VidDepthAny aligns every 32 frames without knowledge of previous frames. The standard deviation rapidly increases with the number of frames, showing it is unable to maintain long-range temporal consistency.  

\section{Ablation Study on Temporal Modules}
We perform an ablation study of the recurrent module in FlashDepth. Specifically, in addition to our original Mamba design, we evaluate xLSTM~\cite{beck:24xlstm} and a transformer-based recurrent network, following CUT3R~\cite{cut3r}. We show results in \cref{tab:ablation-rnn}.

Across datasets, the accuracies are close, with Mamba achieving the highest scores the most times. Mamba is also slightly faster. This shows that the specific choice of recurrent network is not crucial to our method; one could swap out Mamba for future faster or more task-specific options. 

While FlashDepth builds on existing components like Mamba and pretrained depth models, its technical novelty lies in the carefully designed integration of these modules to tackle the challenging setting of online, real-time, high-resolution depth estimation.

\begin{table}[t]
    \centering
    \small
    \setlength{\tabcolsep}{3pt}  
    \renewcommand{\arraystretch}{1.0}  
    \begin{tabular}{lccccc}  
        \toprule
        Dataset / & {ETH3D} & {Sintel} & {Waymo} & {Unreal4K} & {UrbanSyn} \\
        Method & $\delta_1$ ↑ & $\delta_1$ ↑ & $\delta_1$ ↑ &  $\delta_1$ ↑ & $\delta_1$ ↑  \\
        \midrule
        Mamba & \textbf{0.875} &  \textbf{0.642} & \textbf{0.924} & 0.566 & \textbf{0.882} \\
        Transformer & 0.864 & 0.615 & 0.917 & \textbf{0.608} & 0.881\\
        xLSTM & 0.853 & 0.635 & 0.909 & 0.593 & 0.853\\
        \bottomrule
    \end{tabular}
    \caption{Analysis of different recurrent networks. All experiments were run using FlashDepth-L.}
    \label{tab:ablation-rnn}
    \vspace{-1em}
\end{table}

\section{Evaluation Details}

\subsection{F1 Score for Boundary Sharpness}
We follow the F1 metric proposed in Depth Pro~\cite{depth_pro}.
Here, we provide the definition taken from the paper:

We use the pairwise depth ratio of neighboring pixels to define a foreground/background relationship.
Let $i,j$ be the locations of two neighboring pixels. We then define an occluding contour $c_d$ derived from a depth map $d$ as $c_d(i,j) = \left[ \tfrac{d(j)}{d(i)} > (1 + \tfrac{t}{100}) \right]$, where $[\cdot]$ is the Iverson bracket. Intuitively, this indicates the presence of an occluding contour between pixels $i$ and $j$ if their corresponding depth differs by more than $t\%$.
For all pairs of neighboring pixels, we can then compute the precision ($P$) and recall ($R$) as
\begin{gather*}
    \text{P}(t) = \frac{\sum_{i,j\in N(i)} c_d(i,j) \wedge c_{\hat{d}}(i,j)}{ \sum_{i,j\in N(i)} c_{d}(i,j)}, \\
    \text{R}(t) = \frac{\sum_{i,j\in N(i)} c_d(i,j) \wedge c_{\hat{d}}(i,j)}{ \sum_{i,j\in N(i)} c_{\hat{d}}(i,j)}.\label{eq:recall}
\end{gather*}
Note that both $P$ and $R$ are scale-invariant. In our experiments, we report the F1 score.
To account for multiple relative depth ratios,
we further perform a weighted averaging of the F1 values with 10 thresholds that range linearly from $t_{min}=5$ to $t_{max}=25$,
with stronger weights towards high threshold values.
This metric does not require any manual edge annotation,
but simply pixelwise ground truth, which is easily obtained for synthetic datasets.

\subsection{Measuring FPS in Wall Time}
All FPS numbers were reported using the same machine on a single A100 GPU, with float16 precision. We preprocess all 1000 images to the appropriate resolutions and load them onto GPU memory before starting inference. We start the timer after a single warmup iteration: $\verb|torch.cuda.synchronize;start=time.time()|$ and we stop the timer only after all frames have been processed and aligned as needed: $\verb|torch.cuda.synchronize;end=time.time()|$.

We note that our FPS for VidDepthAny is different from that reported on its official repo. 
We were able to match their FPS of 71 on a tensor of shape 1 $\times$ 32 $\times$ 518 $\times$ 518, but we recorded 24 FPS across the entire sequence of shape 1 $\times$ 1000 $\times$ 924 $\times$ 518. Reasons for the discrepancy include the increase in resolution and that that VidDepthAny sets an overlap of 10 images between batches to perform alignment, so the model actually has to process 1504 frames through a sliding window, despite a fixed video length of 1000.

\end{document}